\documentclass{elsart}
\usepackage{amsmath}
\usepackage{mathrsfs}
\usepackage{bbm}
\usepackage{graphicx}
\usepackage{calligra}
\usepackage{quotmark}
\usepackage{longtable}
\usepackage{cases}
\usepackage{lipsum}
\makeatletter
\def\ps@pprintTitle{%
 \let\@oddhead\@empty
 \let\@evenhead\@empty
 \def\@oddfoot{}%
 \let\@evenfoot\@oddfoot}
\makeatother
\begin{document}
\begin{center}
{\Large{\bf Estimating Probability Distributions using ``Dirac'' Kernels (via Rademacher-Walsh Polynomial Basis Functions)  }.}
\end{center}
\begin{center}
{Hamse Y. Mussa and Avid M. Afzal$^{*}$}
\end{center}
\begin{center}
{\it EMIC Consultancy, 2 Stanley Avenue, Barking IG11 0LE, U.K.\\
$^{*}$ Centre for Molecular Sciences Informatics, Cambridge University, Lensfield Road, Cambridge CB2 1EW, UK}
\end{center}
\begin{center}
{\it mussax021@gmail.com;  maa76@cam.ac.uk}
\end{center}

{\bf Abstract}\\
In many applications (in particular information systems, such as pattern-recognition, machine learning, cheminformatics, bioinformatics to name but a few) the assessment of uncertainty is essential -- i.e., the estimation of the underlying probability distribution function. More often than not, the form of this function is unknown and it becomes necessary to non-parametrically construct/estimate it from a given sample.  

One of the methods of choice to non-parametrically estimate the unknown probability distribution function for a given random variable (defined on binary space) has been the expansion of the estimation function in Rademacher-Walsh Polynomial basis functions. 

In this paper we demonstrate that the expansion of the probability distribution function estimation in Rademacher-Walsh Polynomial basis functions is equivalent to the expansion of the function estimation in a set of ``Dirac kernel'' functions. The latter approach can ameliorate the computational bottle-neck and notational awkwardness often associated with the Rademacher-Walsh Polynomial basis functions approach, in particular when the binary input space is large. 
\section{Introduction}
The assessment of uncertainty is important in quantitative science. This requires the estimation of the underlying probability distribution function explicitly or implicitly. However, the form of the function is usually unknown and it becomes necessary to non-parametrically construct/estimate the function from a given sample. When the random variable of interest is an $L$--dimensional binary ``vector' (i.e., it resides in a binary space $\cal{B}$ = $\{0,1 \}^{L}$ ), its {\it L}-dimensional probability distribution function $p({\bf x})$ is often non-parametrically estimated through $2^{L}$ Rademacher-Walsh Polynomial basis functions $\varphi_{i}$ \footnote{According to Duda and Hart \cite{Duda} this basis function set $\{ \varphi_{i}({\bf x})\}^{2^{L}-1}_{i=0}$ consists of a set of polynomials that can be generated by systematically forming the products of the distinct terms $2x_{l}-1$ taken none at a time, one at a time, two at a time, three at a time, and so on, where ${\bf x} = (x_{1}, x_{2},..., x_l, ..., x_{L})$. The resultant set is a complete set satisfying an orthogonality relation in their order -- {\it i.e.}, $\varphi_{i}({\bf x})$ and $\varphi_{k}({\bf x})$ -- with respect to the weighting function $w({\bf x}) = 1$,
\begin{numcases}
{\sum_{\bf x}\varphi_{i}({\bf x})\varphi_{k}({\bf x})=}\nonumber
2^{L}~~~~~~~ &$i= k$\\\nonumber
0~~~~~~~ &$i\neq k$\nonumber
\end{numcases}
where the summation is taken over all $2^{L}$ values of the binary ``vectors''.} \cite{Duda, Hand} as 
\begin{equation}
p({\bf x}) =  \sum^{2^{L}-1}_{i=0}\alpha_{i}\varphi_{i}({\bf x})
\end{equation}
where 
\begin{equation}
\alpha_{i} =  \frac{1}{2^{L}}\sum_{{\bf x}\in{\cal B}}p({\bf x})\varphi_{i}({\bf x})
\end{equation}
The coefficients $\alpha_{i}$ can be estimated as \cite{Duda} 
\begin{equation}
\hat{\alpha}_{i} = \frac{1}{N}\sum^{N}_{j=1}\frac{1}{2^{L}}\varphi_{i}({\bf x}_{j})
\end{equation}
where {\it N} refers to the number of available prototype patterns ${\bf x}_{j}$.  Putting Eq. 3 into Eq. 1 yields \cite{Meisel}   
\begin{eqnarray}\nonumber
\hat{p}({\bf x}) &=& \sum^{2^{L}-1}_{i=0}\frac{1}{N}\sum^{N}_{j=1}\frac{1}{2^{L}}\varphi_{i}({\bf x}_{j})\varphi_{i}({\bf x})\\\nonumber
&=& \frac{1}{N}\sum^{N}_{j=1}\sum^{2^{L}-1}_{i=0}\frac{\varphi_{i}({\bf x}_{j})}{\sqrt{2^{L}}}\frac{\varphi_{i}({\bf x})}{\sqrt{2^{L}}}\\
&=& \frac{1}{N }\sum^{N}_{j=1}K({\bf x}_{j},{\bf x})
\end{eqnarray}  
where 
\begin{equation}
K({\bf x}_{j},{\bf x}) = \displaystyle{\sum^{2^{L}-1}_{i=0}\frac{\varphi_{i}({\bf x}_{j})}{\sqrt{2^{L}}}\frac{\varphi_{i}({\bf x})}{\sqrt{2^{L}}}}
\end{equation}
For all practical purposes $L<<\infty$; besides $\varphi_{i}({\bf x}_{j})$ (and $\varphi_{i}({\bf x})$) can only take values 1, or -1 as illustrated in \cite{Duda, Hand}. And according to \cite{Aronszajn, Shawe--Taylor}, $K({\bf x}_{j},{\bf x})$ can be considered as a valid positive definite kernel function. 

The estimation of $p({\bf x})$ at {\bf x} can be instructively viewed as an average of how similar {\bf x} is to the given $N$ prototype patterns {\bf x}$_{j}$, where $K({\bf x}_{j},{\bf x})$ is the similarity function \cite{Duda, Hand, Meisel,Parzen}. If the available $N$ prototype patterns constituting the sample are distinct instances and $N = 2^{L}$, the estimated coefficients $\hat{\alpha}_{i}$ are exact \cite{Tou}. However, exact or not, the expansion in Eq. 1 requires $2^{L}$ Rademacher-Walsh Polynomial basis functions, which can make the estimation notationally clumsy and computationally complicated whenever the value of $L$ is large \cite{Duda,Hamse}. Thus, for Eq. 4 to have any practical use, knowledge of the closed form of the kernel function $K({\bf x}_{j},{\bf x})$ is essential. In the following section we demonstrate that the function $K({\bf x}_{j},{\bf x})$ in Eq. 5 is a ``Dirac'' kernel function \cite{Jacob}. Our concluding remarks are in the final section.
 
 \section{Main Idea}

Here we present the nub of the paper: $K({\bf x}_{j},{\bf x})$ is a ``Dirac'' kernel function. 
 
\textbf{Theorem}~\emph{~If ${\bf x}$ and ${\bf x}_{j}\in{\cal B}$, and $\varphi_{i}(.)$ are Rademacher-Walsh Polynomial basis functions on ${\cal B}$, then 
\begin{numcases}
{K({\bf x}_{j},{\bf x})=\sum^{2^{L}-1}_{i=0}\frac{\varphi_{i}({\bf x}_{j})}{\sqrt{2^{L}}}\frac{\varphi_{i}({\bf x})}{\sqrt{2^{L}}}=}\nonumber
1&$~{\bf x}_{j}={\bf x}$\\
0&$~{\bf x}_{j}\neq{\bf x}$
\end{numcases} 
 i.e., $K({\bf x}_{j},{\bf x})$ is a ``Dirac kernel'' function.}\\
where {\bf x}$_{j} = {\bf x}$ means that $x_{j1} = x_{1}, x_{j2} = x_{2}, ..., x_{jL} = x_{L}$, with x$_{jl}$ and x$_{l}$ referring to the binary--valued $l^{th}$ elements of {\bf x}$_{j}$ and {\bf x}, respectively. 
  
As described in the Introduction, the set $\{\varphi_{i}({\bf x})\}^{2^{L}-1}_{i=0}$ is obtained by systematically forming products of $(2x_{l} -1)$ none at a time, one at a time, two at a time, three at a time, {\it etc.}, where $l =1, 2, ..., L$. By the same token the set $\{\varphi_{i}({\bf x}_j)\varphi_{i}({\bf x})\}^{2^{L}-1}_{i=0}$ is obtained by forming products of the distinct terms $(2x_{jl} -1)(2x_{l} -1)$ none at a time, one at a time, two at a time, three at a time, and so on.

\textbf{Lemma 1}~\emph{Let $a_{1}, a_{2},...,a_{L}$ be {\it L} distinguishable real variables which can take the values of 1 and -1, and that their combinatorial compositions can be considered as products.The sum of their possible combinatorial compositions $z_{i}$, with $i =0, 1, ..., 2^{L}-1$ is
\begin{numcases}
{\sum^{2^L-1}_{i=0}z_{i}=}
2^{L}&$if~a_1, a_2, ..., a_L= 1$\\\nonumber
0&$if~not$\nonumber
\end{numcases}  
} 

\emph{ Proof:}

The possible combinations are the $L$ variables chosen: no variable; 1 variable, $a_{i}$, at a time; 2 variables, $a_{i}a_{j}$, at a time; three variables,$a_{i}a_{j}a_{k}$, at a time;,...,; or $L$ variables, $a_{1}a_{2}...a_{L}$, at a time.  

~~~~\emph{If all the $L$ variables are positive(Scenario1), i.e., $a_{k}$ =  +1 (where $k = 1, 2, .., L$),   then $z_{0} = +1$ (when no variable is chosen); $z_1= a_{1} = +1, z_2 = a_{2} = +1, ...,  z_L = a_{L} = +1; z_{L+1} = a_{1}a_{2} = +1, z_{L+2} = a_{1}a_{3} = +1,...,z_{L+\frac{L(L-1)}{2}} = a_{L-1}a_{L} = +1$;, ...,; and $z_{2^{L}-1}= a_{1}a_{2}...a_{L} = 1$}. 

Self-evidently the number of times that none of the variables is chosen is ${^L}C_{0}={L \choose 0}$; the number of combinatorial terms containing one variable is ${^L}C_{1} = {L \choose 1}$; and the number combinatorial terms consisting of two, three, four, ..., and $L$ variables are ${^L}C_{\rho} = {L \choose \rho}$, $\rho$ being 2, 3, ..., and $L$, respectively. This means
\begin{equation}
\displaystyle\sum^{2^{L}-1}_{i=0}z_{i} = \displaystyle\sum^{L}_{\varrho=0}{L \choose \varrho}=2^{L}
\end{equation}
 
~~~~\emph{ If all the $L$ variables take the value of -1 (Scenario2), i.e., $a_{k}$=  -1 where $k$ is as defined before, then $z_{0} = +1$; $z_1= a_{1} = (-1)^{1}, z_2 = a_{2} = (-1)^{1},..., z_L = a_{L} = (-1)^{1}; z_{L+1} = a_{1}a_{2} = (-1)^{2}, z_{L+2} = a_{1}a_{3} = (-1)^{2},...,z_{L+\frac{L(L-1)}{2}} = a_{L-1}a_{L} = (-1)^{2}$;,...,; and $z_{2^{L}-1}= a_{1}a_{2}...a_{L} =  (-1)^{L}$}. By the same token (as we reasoned above):  ${^L}C_{\varrho} = (-1)^{\varrho}{L \choose \varrho}$ with $\varrho = 0, 1, 2, ...,L$. Here
\begin{equation}
\displaystyle\sum^{2^{L}-1}_{i=0}z_{i} =\displaystyle\sum^{L}_{\varrho=0}(-1)^{\varrho}{L \choose \varrho}, 
\end{equation}
where obviously $\sum^{L}_{\varrho=0}(-1)^{\varrho}{L \choose \varrho}=0$.

In the final scenario (\emph{Scenario3}):~\emph{For no specific reason, let us consider that $m$ and $k$ denote the number of variables that take the values -1 and 1, respectively, where $L = m + k$.}  In this scenario
\begin{eqnarray}
\displaystyle\sum^{2^{L}-1}_{i=0} z_{i} = \displaystyle\sum^{m+k}_{\varrho=0}{^{m+k}}C_{\varrho}
\end{eqnarray}
 It can readily be shown by induction that $ \displaystyle\sum^{m+k}_{\varrho=0}{^{m+k}}C_{\varrho}=0$ if one makes use of these three identities: 
\\
I:~${^n}C_{r} = {^{n-1}}C_{r} + {^{n-1}}C_{r-1}$,\\
II:~${^{n+r}}C_{n+r} = {^{n+r-1}}C_{n+r-1}$, and\\ 
III:~${^{n+j}}C_{0} = {^{n+j-1}}C_{0}$,\\
whereby \emph{r}, \emph{j}, \emph{n} are non-negative integers and \emph{r}$\leq$\emph{n} \cite{Riley}.

With {\it k = 1}, {\it i.e.}, $\displaystyle\sum^{m+k}_{\varrho=0}{^{m+k}}C_{\varrho}$ becomes  $\displaystyle\sum^{m+1}_{\varrho=0}{^{m+1}}C_{\varrho}$, which can be expressed as
\begin{eqnarray}\nonumber 
\displaystyle\sum^{m+1}_{\varrho=0}{^{m+1}}C_{\varrho}&=& {^{m+1}}C_{0} + \displaystyle\sum^{m}_{\varrho=1}{^{m+1}}C_{\varrho}+ {^{m+1}}C_{m+1}\\\nonumber
\end{eqnarray}
Making use of Identity I, the ${^{m+1}}C_{\varrho}$ on the RHS of the equation above becomes ${^{m}}C_{\varrho}+ {^{m}}C_{\varrho -1}$, {\it i.e.}, the equation can be rewritten as
\begin{eqnarray}\nonumber 
\displaystyle\sum^{m+1}_{\varrho=0}{^{m+1}}C_{\varrho}&=&{^{m+1}}C_{0} + \displaystyle\sum^{m}_{\varrho=1}{^{m}}C_{\varrho}+ \displaystyle\sum^{m}_{\varrho=1}{^{m}}C_{\varrho -1}+ {^{m+1}}C_{m+1}\nonumber,
\end{eqnarray}
which can be modified further by applying Identities III and II to the first and last terms on its RHS, respectively, resulting in
\begin{eqnarray}\nonumber 
\displaystyle\sum^{m+1}_{\varrho=0}{^{m+1}}C_{\varrho}&=&{^{m}}C_{0} + \displaystyle\sum^{m}_{\varrho=1}{^{m}}C_{\varrho}+ \displaystyle\sum^{m}_{\varrho=1}{^{m}}C_{\varrho -1}+ {^{m}}C_{m}=2\displaystyle\sum^{m}_{\varrho=0}{^{m}}C_{\varrho}\nonumber
\end{eqnarray}
In \emph{Scenario2} we have demonstrated that in the case that all the variables (denoted here by {\it m}) take the value of -1,  ${^{m}}C_{\varrho}= (-1)^{\varrho}{m \choose \varrho}$. This means
\begin{equation} 
\displaystyle\sum^{m+1}_{\varrho=0}{^{m+1}}C_{\varrho}= 2\displaystyle\sum^{m}_{\varrho=0}(-1)^{\varrho}{m \choose \varrho}
\end{equation}

In the case of  {\it k =2}, $\displaystyle\sum^{m+k}_{\varrho=0}{^{m+k}}C_{\varrho}$ becomes $\displaystyle\sum^{m+2}_{\varrho=0}{^{m+2}}C_{\varrho}$, which can be expressed as  
\begin{eqnarray}\nonumber 
\displaystyle\sum^{m+2}_{\varrho=0}{^{m+2}}C_{\varrho}&=& {^{m+2}}C_{0} + \displaystyle\sum^{m+1}_{\varrho=1}{^{m+2}}C_{\varrho}+ {^{m+2}}C_{m+2}\\\nonumber
\end{eqnarray}
 Applying Identity I to $^{{m+2}}C_{\varrho}$ in the middle term on the RHS of the equation above, we obtain
\begin{equation}\nonumber 
\displaystyle\sum^{m+2}_{\varrho=0}{^{m+2}}C_{\varrho}={^{m+2}}C_{0} + \displaystyle\sum^{m+1}_{\varrho=1}{^{m+1}}C_{\varrho}+ \displaystyle\sum^{m+1}_{\varrho=1}{^{m+1}}C_{\varrho -1}+ {^{m+2}}C_{m+2}\nonumber
\end{equation}
By following the same line of reasoning as employed in the case of {\it k}=1 and applying Identities III and II to the first and last terms on the RHS of the equation above, respectively, gives
\begin{equation}
\displaystyle\sum^{m+2}_{\varrho=0}{^{m+2}}C_{\varrho}=2\displaystyle\sum^{m+1}_{\varrho=0}{^{m+1}}C_{\varrho},
\end{equation}
whereby $2\displaystyle\sum^{m+1}_{\varrho=0}{^{m+1}}C_{\varrho}$ can be expressed as
\begin{eqnarray}
2\displaystyle\sum^{m+1}_{\varrho=0}{^{m+1}}C_{\varrho} = 2\bigg[2 \displaystyle\sum^{m}_{\varrho=0}(-1)^{\varrho}{m \choose \varrho}\bigg]=2^{2}\displaystyle\sum^{m}_{\varrho=0}(-1)^{\varrho}{m \choose \varrho}
\end{eqnarray}

 For $\sum^{m+k}_{\varrho=0}{^{m+k}}C_{\varrho}$, one just needs to repeat the process above {\it k} times, which gives
\begin{eqnarray}\nonumber 
\displaystyle\sum^{m+k}_{\varrho=0}{^{m+k}}C_{\varrho}=2^{k}\displaystyle\sum^{m}_{\varrho=0}(-1)^{\varrho}{m \choose \varrho}
\end{eqnarray} 
 
In\emph{ Scenario2}, where $m=L$, it was shown that $\displaystyle\sum^{m}_{\varrho=0}(-1)^{\varrho}{m \choose \varrho} = 0$. Thus
\begin{eqnarray}\nonumber 
\displaystyle\sum^{m+k}_{\varrho=0}{^{m+k}}C_{\varrho}=2^{k}\displaystyle\sum^{m}_{\varrho=0}(-1)^{\varrho}{m \choose \varrho} = 0
\end{eqnarray}  
I.e.
\begin{equation}
\displaystyle\sum^{2^{L}-1}_{i=0}z_{i} = \displaystyle\sum^{m+k}_{\varrho=0}{^{m+k}}C_{\varrho} =0
\end{equation}  
This finalizes the proof of \textbf{Lemma 1}.

\section{\emph{Proof of} \textbf{Theorem 1}}

As described above, $\varphi_{0}({\bf x}_j)\varphi_{0}({\bf x}) = 1$ and the terms $\varphi_{i}({\bf x}_j)\varphi_{i}({\bf x})$ take the values +1 or -1, where $i = 1, 2,..., L$. 

Now, if we consider $\varphi_{1}({\bf x}_j)\varphi_{1}({\bf x})$, $\varphi_{2}({\bf x}_j)\varphi_{2}({\bf x})$, ..., and $\varphi_{L}({\bf x}_j)\varphi_{L}({\bf x})$ as the real $L$ variables in \textbf{ Lemma 1}, then \\
\\
$z_{0} = \varphi_{0}({\bf x}_j)\varphi_{0}({\bf x})$,\\
$z_{1} = \varphi_{1}({\bf x}_j)\varphi_{1}({\bf x})$, \\
. \\
.\\
.\\
$z_{2^{L}-1} = \varphi_{2^{L}-1}({\bf x}_j)\varphi_{2^{L}-1}({\bf x}) = [\varphi_{1}({\bf x}_j)\varphi_{1}({\bf x})][\varphi_{2}({\bf x}_j)\varphi_{2}({\bf x})]...[\varphi_{L}({\bf x}_j)\varphi_{L}({\bf x})]$.   

Then by the virtue of \textbf{Lemma 1}, 
 
\begin{numcases}
{\displaystyle\sum^{2^{L}-1}_{i=0}z_{i} = \sum^{2^{L}-1}_{i=0}\varphi_{i}({\bf x}_{j})\varphi_{i}({\bf x})=}\nonumber
2^{L}&$if~\varphi_{1}({\bf x}_{j})\varphi_{1}({\bf x}), ..., \varphi_{L}({\bf x}_{j})\varphi_{L}({\bf x}) = 1$\\
0&$if~not$
\end{numcases} 
Recall that the elements of the set $\{\varphi_{i}({\bf x}_{j})\varphi_{i}({\bf x})\}^{L}_{i=1}$ take the value of 1 only if {\bf x} = {\bf x}$_{j}$. Multiplying on both side of Eq. 15 by $\frac{1}{\sqrt{2^{L}}}\frac{1}{\sqrt{2^{L}}}$ yields

\begin{numcases}
{\sum^{2^{L}-1}_{i=0}\frac{\varphi_{i}({\bf x}_{j})}{\sqrt{2^{L}}}\frac{\varphi_{i}({\bf x})}{\sqrt{2^{L}}}=}
1&$if~{\bf x}_{j}={\bf x}$\\\nonumber
0&$if~{\bf x}_{j}\neq{\bf x}$\nonumber
\end{numcases} 
which is Eq. 6 and this completes the proof of \textbf{Theorem 1}.

\section{Conclusion}

In this paper we have demonstrated that, on binary space ${\cal B}$, the expansion of the probability distribution estimation function in Rademacher-Walsh Polynomial basis functions is equivalent to the expansion of the estimation function in a set of Dirac kernel functions. The probability distribution estimation based on the Dirac kernel function scheme certainly alleviates both the computational bottle-necks and notational complexity associated with the Rademacher-Walsh Polynomial basis function approach, in particular when ${\cal B}$ is large. 

{\bf Acknowledgements}
It is a great pleasure to acknowledge Dr. J. B. O. Mitchell for reading the manuscript and his useful comments.

\end{document}